\documentclass[10pt,twocolumn,letterpaper]{article}

\usepackage{cvpr}
\usepackage{times}
\usepackage{epsfig}
\usepackage{graphicx}
\usepackage{amsmath}
\usepackage{amssymb}

\usepackage{url}            
\usepackage{booktabs}       
\usepackage{amsfonts}       
\usepackage{nicefrac}       
\usepackage{microtype}      
\usepackage{textcomp}
\usepackage{xcolor}
\usepackage{multirow}

\newcommand\abbr[1]{\textsc{#1}}

\usepackage{caption}
\usepackage{subcaption}
\usepackage{paralist}
\usepackage{array}
\usepackage{placeins}
\usepackage{epstopdf}
\usepackage[titletoc,title]{appendix}

\usepackage[ruled]{algorithm2e}
\makeatletter
\def\BState{\State\hskip-\ALG@thistlm}
\makeatother

\ifcvprfinal\fi

\usepackage{color}

\usepackage{enumitem}
\setitemize{noitemsep,topsep=0pt,parsep=0pt,partopsep=0pt}

\newcommand\blfootnote[1]{%
  \begingroup
  \renewcommand\thefootnote{}\footnote{#1}%
  \addtocounter{footnote}{-1}%
  \endgroup
}

\usepackage[pagebackref=true,breaklinks=true,letterpaper=true,colorlinks,bookmarks=false]{hyperref}

\cvprfinalcopy 

\def\cvprPaperID{1296} 

\ifcvprfinal\fi
\begin{document}

\title{Dynamic Few-Shot Visual Learning without Forgetting}

\author{
Spyros Gidaris\\
University Paris-Est, LIGM\\
Ecole des Ponts ParisTech\\
{\tt\small spyros.gidaris@enpc.fr}
\and
Nikos Komodakis\\
University Paris-Est, LIGM\\
Ecole des Ponts ParisTech\\
{\tt\small nikos.komodakis@enpc.fr}\\
}

\maketitle
\thispagestyle{empty}

\begin{abstract}
The human visual system has the remarkably ability to be able to effortlessly learn novel concepts from only a few examples. 
Mimicking the same behavior on machine learning vision systems is an interesting and very challenging research problem with many practical advantages on real world vision applications. In this context, the goal of our work is to devise a few-shot visual learning system that during 
test time it will be able to efficiently learn novel categories from only a few training data while at the same time it will not forget the initial categories on which it was trained (here called base categories).
To achieve that goal we propose (a) to extend an object recognition system with an attention based few-shot classification weight generator,
and (b) to redesign the classifier of a ConvNet model as the cosine similarity function between feature representations and classification weight vectors. 
The latter, apart from unifying the recognition of both novel and base categories,
it also leads to feature representations that generalize better on ``unseen" categories.
We extensively evaluate our approach on Mini-ImageNet where we manage to improve the prior state-of-the-art on few-shot recognition 
(i.e., we achieve $56.20\%$ and $73.00\%$ on the 1-shot and 5-shot settings respectively) 
while at the same time we do not sacrifice any accuracy on the base categories, which is a characteristic that most prior approaches lack.
Finally, we apply our approach on the recently introduced few-shot benchmark of Bharath and Girshick~\cite{hariharan2016low} where we also achieve state-of-the-art results.
The code and models of our paper will be published on:
\url{https://github.com/gidariss/FewShotWithoutForgetting}.
\blfootnote{This work was supported by the ANR SEMAPOLIS project, an INTEL gift, and hardware donation by NVIDIA.}
\end{abstract}

\section{Introduction}

Over the last few years, deep convolutional neural networks~\cite{krizhevsky2012imagenet, simonyan2014very, szegedy2015going, he2016deep} (ConvNets) have achieved impressive results on image classification tasks, such as object recognition~\cite{russakovsky2015imagenet} or scene classification~\cite{NIPS2014_5349}.
In order for a ConvNet to successfully learn to recognize a set of visual categories (e.g., object categories or scene types), it requires to manually collect and label thousands of training examples per target category and to apply on them an iterative gradient based optimization routine~\cite{lecun1998gradient} that is extremely computationally expensive, 
e.g., it can consume hundreds or even thousands of GPU hours.
Moreover, the set of categories that the ConvNet model can recognize remains fixed after training.
In case we would like to expand the set of categories that the ConvNet can recognize, 
then we need to collect training data for the novel categories (i.e., those that they were not in the initial training set) and restart the aforementioned computationally costly training procedure this time on the enhanced training set such that we will avoid catastrophic interference.
Even more, it is of crucial importance to have enough training data for the novel categories (e.g., thousands of examples per category) otherwise we risk overfitting on them.

\emph{In contrast, the human visual system exhibits the remarkably ability to be able to effortlessly learn novel concepts from only one or a few examples and reliably recognize them later on}. 
It is assumed that the reason the human visual system is so efficient when learning novel concepts is that it exploits its past experiences about the (visual) world. 
For example, a child,
having accumulated enough knowledge about mammal animals and in general the visual world,
can easily learn and generalize the visual concept of ``rhinoceros" from only a single image.
Mimicking that behavior on artificial vision systems is an interesting and very challenging research problem with many practical advantages,
such as developing real-time interactive vision applications for portable devices (e.g., cell-phones). 

Research on this subject is usually termed \emph{few-shot learning}.
However, most prior methods neglect to fulfill two very important requirements for a good few-shot learning system:
\textbf{(a)} the learning of the novel categories needs to be fast, and 
\textbf{(b)} to not sacrifice any recognition accuracy on the initial categories that the ConvNet was trained on, i.e., to not ``forget" (from now on we will refer to those initial categories by calling them base categories). 
Motivated by this observation, in this work we propose to tackle the problem of few-shot learning under a more realistic setting, where a large set of training data is assumed to exist for a set of base categories and, using these data as the sole input, we want to develop an object recognition learning system that, not only is able to recognize these base categories, but also learns to dynamically recognize novel categories from only a few training examples (provided only at test time) while also not forgetting the base ones or requiring to be re-trained on them (\emph{dynamic few-shot learning without forgetting}).  
Compared to prior approaches, 
we believe that this setting more closely resembles the human visual system behavior (w.r.t. how it learns novel concepts). In order to achieve our goal, we propose two technical novelties.

\textbf{Few-shot classification-weight generator based on attention.}
A typical ConvNet based recognition model,
in order to classify an image,
first extracts a high level feature representation from it and then 
computes per category classification scores by applying a set of classification weight vectors (one per category) to the feature.
Therefore, in order to be able to recognize novel categories we must be able to generate classification weight vectors for them. 
In this context,
the first technical novelty of our work is that we enhance 
a typical object recognition system with an extra component, 
called \emph{few-shot classification weight generator} that 
accepts as input 
a few training examples of a novel category (e.g., no more than five examples) and, based on them, generates a classification weight vector for that novel category.
Its key characteristic is that in order to compose novel classification weight vectors, 
it explicitly exploits the acquired past knowledge about the visual world by incorporating an attention mechanism over the classification weight vectors of the base categories.
This attention mechanism offers a significant boost on the recognition performance of novel categories, especially when there is only a single training example available for
learning them. 

\textbf{Cosine-similarity based ConvNet recognition model.} 
In order for the \emph{few-shot classification weight generator} to be successfully incorporated into the rest of the recognition system, it is essential
the
ConvNet model to be able to simultaneously handle the classification weight vectors of both base and novel categories.
However, as we will explain in the methodology, this is not feasible with the typical dot-product based classifier (i.e., the last linear layer of a classification neural network).
Therefore, in order to overcome this serious issue,
our second technical novelty is 
to implement the classifier as a cosine similarity function between the feature representations and the classification weight vectors.
Apart from unifying the recognition of both base and novel categories,
features learned with the cosine-similarity based classifier turn out to generalize significantly better on novel categories than those learned with a dot-product based classifier.
Moreover, we demonstrate in the experimental section that,
by simply training a cosine-similarity based ConvNet recognition model, 
we are able to learn feature extractors that when used for image matching they surpass prior state-of-the-art approaches on the few-shot recognition task.

To sum up, our contributions are:
\textbf{(1)}
We propose a few-shot object recognition system that is capable of dynamically learning novel categories from only a few training data while at the same time does not forget the base categories on which it was trained. 
\textbf{(2)}
In order to achieve that we introduced two technical novelties, an attention based few-shot classification weight generator, and to 
implement the classifier of a ConvNet model as a cosine similarity function between feature representations and classification vectors. 
\textbf{(3)}
We extensively evaluate our object recognition system on Mini-ImageNet, both w.r.t. its few-shot object recognition performance and its ability to not forget the base categories, and we report state-of-the-art results that surpass prior approaches by a very significant margin.
\textbf{(4)}
Finally, we apply our approach on the recently introduced fews-shot benchmark of Bharath and Girshick~\cite{hariharan2016low} where we achieve state-of-the-art results.

In the following sections, 
we provide related work in~\S\ref{sec:relatedwork}, 
we describe our few-shot object learning methodology in~\S\ref{sec:methodology},
we provide experimental results in~\S\ref{sec:experimentalresults}, and finally 
we conclude in~\S\ref{sec:conclusion}.

\section{Related work} \label{sec:relatedwork} 

Recently, there is resurgence of interest on the few-shot learning problem. In the following we briefly discuss the most relevant approaches to our work.

\textbf{Meta-learning based approaches.}
Meta-learning approaches typical involve a meta-learner model that given a few training examples of a new task it tries to quickly learn a learner model that ``solves" this new task~\cite{schmidhuber1997shifting,thrun1998lifelong,andrychowicz2016learning,munkhdalai2017meta,santoro2016one}. 
Specifically, Ravi and Larochelle~\cite{ravi2016optimization} propose a LSTM~\cite{hochreiter1997long} based meta-learner that is trained 
given as input a few training examples of a new classification task to sequentially generate parameter updates that will optimize the classification performance of a learner model on that task. Their LSTM also learns the parameter initialization of the learner model.
Finn \etal~\cite{finn2017model} simplified the above meta-learner model and only learn the initial learner parameters such that only a few gradient descent steps w.r.t. those initial parameters will achieve the maximal possible performance on the new task.
Mishra \etal~\cite{mishra2017meta} instead propose a generic temporal convolutional network that given as input 
a sequence of a few labeled training examples and then an unlabeled test example, it predicts the label of that test example. 
Our system also includes a meta-learner network component, 
the few-shot classification weight generator.

\textbf{Metric-learning based approaches.}
In general, metric learning approaches attempt to learn feature representations that preserve the class neighborhood structure (i.e., features of the same object are closer than features of different objects).
Specifically, Koch \etal~\cite{koch2015siamese} formulated the one-shot object recognition task as image matching and train Siamese neural networks to compute the similarity between a training example of a novel category and a test example. 
Vinyals \etal~\cite{vinyals2016matching} proposed Matching Networks that in order to classify a test example it employs 
a differentiable nearest neighbor classifier implemented with an attention mechanism over the learned representations of the training examples.
Prototypical Networks~\cite{snell2017prototypical} learn to classify test examples by computing distances to prototype feature vectors of the novel categories.  
They propose to learn the prototype feature vector of a novel category as the average of the feature vectors extracted by the training examples of that category.
A similar approach was proposed before by Mensink \etal~\cite{mensink2012metric}
and Prototypical Networks can be viewed as an adaption of that work for ConvNets.
Despite their simplicity, Prototypical Networks demonstrated state-of-the-art performance.
Our few-shot classification weight generator also includes a feature averaging mechanism. However, more than that, it also explicitly exploits past knowledge about the visual world with an attention based mechanism and the overall framework allows to perform unified recognition of both base and novel categories without altering the way base categories are learnt and recognized.

In a different line of work, 
Bharath and Girshick~\cite{hariharan2016low} 
propose to use during training a $l_2$ regularization loss on the feature representations that forces them to better generalize on ``unseen" categories. In our case, the cosine-similarity based classifier, apart from unifying the recognition of both base and novel categories, it also leads to feature representations that are able to better generalize on ``unseen" categories.
Also, their framework is able to recognize both base and novel categories as ours. 
However, to achieve that goal they re-train the classifier on both the base categories (with a large set of training data) and the novel categories (with few training data), which is in general slow and requires constantly maintaining in disc a large set of training data. 

\section{Methodology} \label{sec:methodology}
\begin{figure*}
\renewcommand{\figurename}{Figure}
\renewcommand{\captionlabelfont}{\bf}
\renewcommand{\captionfont}{\small} 
\begin{center}
\includegraphics[width=0.75\textwidth]{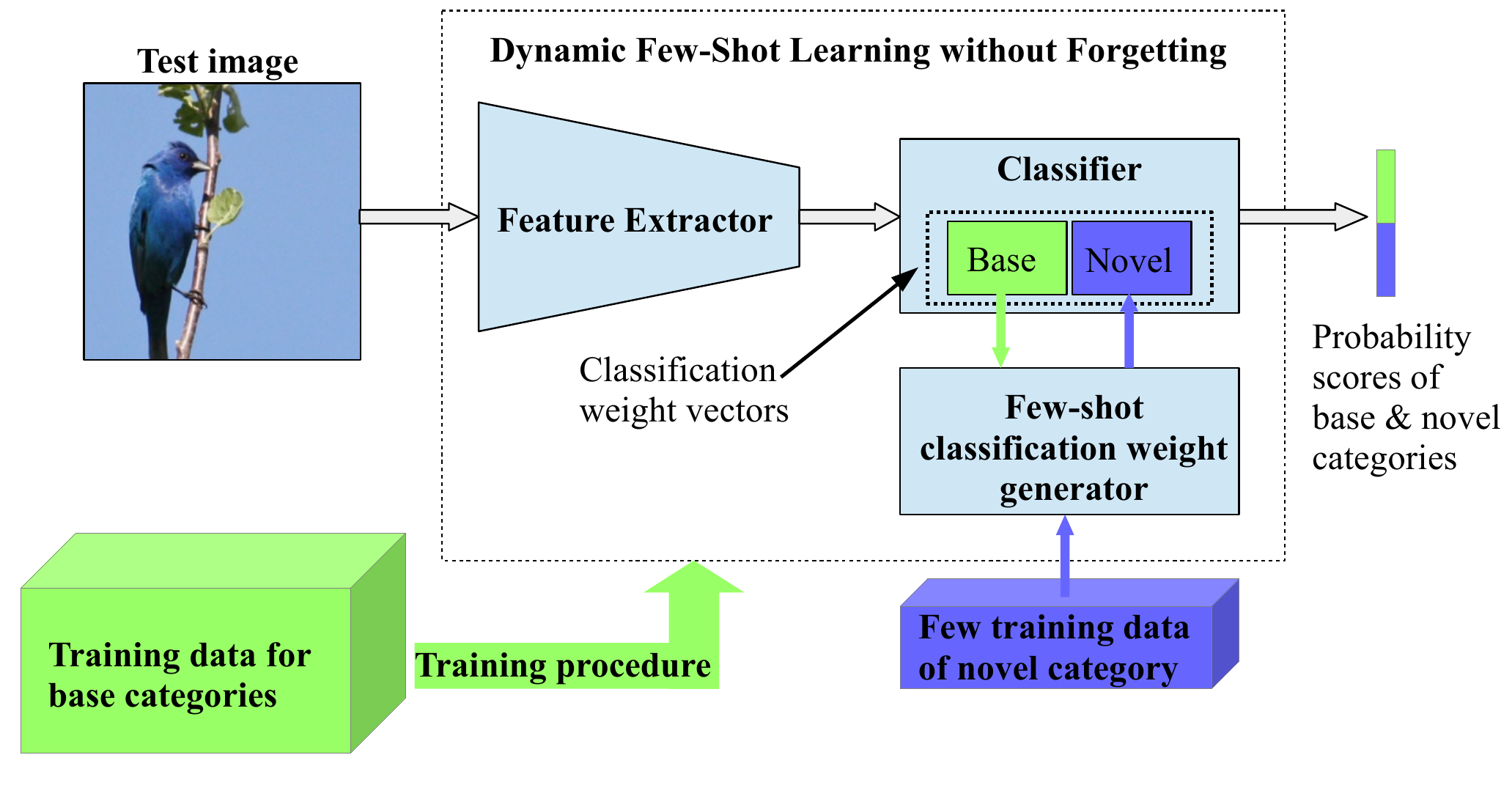}
\end{center}
\vspace{-15pt}
\caption{\small{
Overview of our system.
It consists of: 
(a) a \emph{ConvNet based recognition model} (that includes a feature extractor and a classifier) and (b) a \emph{few-shot classification weight generator}. 
Both are trained on a set of base categories for which we have available a large set of training data.
During test time, the weight generator gets as input a few training data of a novel category and the classification weight vectors of base categories (green rectangle inside the classifier box) and generates a classification weight vector for this novel category (blue rectangle inside the classifier box).
This allows the ConvNet to recognize both base and novel categories.
}}
\label{fig:overview}
\end{figure*}

As an input to our object recognition learning system we assume that there exists a dataset
of $K_{base}$ base categories:
\begin{equation} \label{eq:dtrain_base}
D_{train} = \bigcup_{b=1}^{K_{base}} \{ x_{b,i} \}_{i=1}^{N_b} \text{ ,}
\end{equation}
where $N_b$ is the number of training examples of the $b$-th category and $x_{b,i}$ is its $i$-th training example. 
Using this as the only input, the goal of our work is to be able to both learn to accurately recognize base categories and to learn to perform few-shot learning of novel categories in a dynamic manner and without forgetting the base ones. 
An overview of our framework is provided in Figure~\ref{fig:overview}. 
It consists of two main components, 
a \emph{ConvNet-based recognition model} that is able to recognize both base and novel categories and 
a \emph{few-shot classification weight generator} that dynamically generates classification weight vectors for the novel categories at test time:

\textbf{ConvNet-based recognition model.}
It consists of \textbf{(a)} a feature extractor $F(. | \theta)$ (with learnable parameters $\theta$) that extracts a $d$-dimensional feature vector $z = F(x | \theta) \in \mathbb{R}^{d}$ from an input image $x$, and \textbf{(b)} a classifier $C(. | W^*)$, where $W^*= \{ w^*_k \in \mathbb{R}^{d} \}_{k=1}^{K^*}$ are a set of $K^*$ classification weight vectors - one per object category, that takes as input the feature representation $z$ and returns a $K^*$-dimensional vector with the probability classification scores $p = C(z|W^*)$ of the $K^*$ categories. 
Note that in a typical convolutional neural network the feature extractor is the part of the network that starts from the first layer and ends at the last hidden layer while the classifier is the last classification layer. During the single training phase of our algorithm, we learn the $\theta$ parameters and the classification weight vectors of the base categories $W_{base} = \{ w_{k} \}_{k=1}^{K_{base}}$ such that by setting $W^* = W_{base}$ the ConvNet model will be able to recognize the base object categories.

\textbf{Few-shot classification weight generator.}
This comprises a meta-learning mechanism that,
during test time, 
takes as input a set of $K_{novel}$ novel categories with few training examples per category
\begin{equation} \label{eq:dtrain_novel}
D_{novel} = \bigcup_{n=1}^{K_{novel}} \{ x'_{n,i} \}_{i=1}^{N'_n} \text{ ,}
\end{equation}
where $N'_n$ is the number of training examples of the $n$-th novel category and $x'_{n,i}$ is its $i$-th training example,
and is able to dynamically assimilate the novel categories on the repertoire of the above ConvNet model.
More specifically, 
for each novel category $n \in [1, N_{novel}]$,
the few-shot classification weight generator $G(.,.|\phi)$
gets as input the feature vectors $Z'_{n} = \{ z'_{n,i} \}_{i=1}^{N'_n}$ of its $N'_n$ training examples,
where $z'_{n,i} = F(x'_{n,i} |\theta)$,
and the classification weight vectors of the base categories $W_{base}$ and generates a classification weight vector $w'_{n} = G(Z'_{n},W_{base}|\phi)$ for that novel category. 
Note that $\phi$ are the learnable parameters of the few-shot weight generator, which are learned during the single training phase of our framework. 
Therefore, if 
$W_{novel} = \{ w'_{n} \}_{n=1}^{K_{novel}}$
are the classification weight vectors of the novel categories inferred by the few-shot weight generator, then by 
setting $W^* = W_{base} \cup W_{novel}$ on the classifier $C(.|W^*)$ 
we enable the ConvNet model to recognize both base and novel categories.

A key characteristic of our framework is that it is able to effortlessly (i.e., quickly during test time) learn novel categories and at the same time recognize both base and novel categories in a unified manner.
In the following subsections, we will describe in more detail the 
ConvNet-based recognition model in~\S\ref{sec:classifier} 
and the few-shot weight generator in~\S\ref{sec:generator}.
Finally, we will explain the training procedure in~\S\ref{sec:training}.

\subsection{Cosine-similarity based recognition model} \label{sec:classifier}

A crucial difference of our ConvNet based recognition model compared to a standard one is that it should be able to dynamically incorporate at test time a variable number of novel categories (through the few-shot weight generator).

The standard setting for classification neural networks is, 
after having extracted the feature vector $z$, 
to estimate the classification probability vector $p = C(z|W^*)$ by first computing the raw classification score $s_k$ of each category $k \in [1, K^*]$ using the dot-product operator:
\begin{equation} \label{eq:basicdot}
s_k = z^\intercal w^*_k \text{ ,}
\end{equation}
where $w_k$ is the $k$-th classification weight vector in $W^*$,
and then applying the softmax operator across all the $K^*$ classification scores, 
i.e., $p_k = softmax(s_j)$,
where $p_k$ is the $k$-th classification probability of $p$.
In our case the classification weight vectors $w^*_k$ could come both from the base categories, i.e., $w^*_k \in W_{base}$, and the novel categories, i.e., $w^*_k \in W_{novel}$.
However, the mechanisms involved during learning those classification weights are very different.
The base classification weights, starting from their initial state, are slowly modified (i.e., slowly learned) with small SGD steps and thus their magnitude changes slowly over the course of their training.
In contrast, the novel classification weights are dynamically predicted (i.e., quickly learned) by the weight generator based on the input training feature vectors and thus their magnitude depends on those input features.
Due to those differences, the weight values in those two cases (i.e., base and novel classification weights) can be completely different, 
and so the same applies to the raw classification scores computed with the dot-product operation, which can thus have 
totally different magnitudes depending on whether they come from the base or the novel categories. 
This can severely impede the training process and, in general, does not allow to have a unified recognition of both type of categories.
In order to overcome this critical issue, 
we propose to modify the classifier $C(.|W^*)$ and compute the raw classification scores using the cosine similarity operator:
\begin{equation} \label{eq:basic}
s_k = \tau \cdot cos(z,w^*_k) = \tau \cdot \overline{z}^\intercal \overline{w}^*_k \text{ ,}
\end{equation}
where $\overline{z} = \frac{z}{\|z\|}$ and $\overline{w}^*_k = \frac{w^*_k}{\|w^*_k\|}$ are the $l_2$-normalized vectors (from now on we will use the overline symbol $\overline{z}$ to indicate that a vector $z$ is $l_2$-normalized), and $\tau$ is a learnable scalar value\footnote{
The scalar parameter $\tau$ is introduced in order to control the peakiness of the probability distribution generated by the softmax operator since the range of the cosine similarity is fixed to $[-1, 1]$. 
In all of our experiments $\tau$ is initialized to 10.}.
Since the cosine similarity can be implemented by first $l_2$-normalizing the feature vector $z$ and the classification weight vector $w^*_k$ and then applying the dot-product operator,
the absolute magnitudes of the classification weight vectors can no longer affect the value of the raw classification score (as a result of the $l_2$ normalization that took place). 

\begin{figure*}[t]
\centering
\renewcommand{\figurename}{Figure}
\renewcommand{\captionlabelfont}{\bf}
\renewcommand{\captionfont}{\small} 
\centering
\begin{subfigure}[b]{\textwidth}
\center
        \begin{center}
        \begin{subfigure}[b]{0.49\textwidth}
        \includegraphics[width=\textwidth]{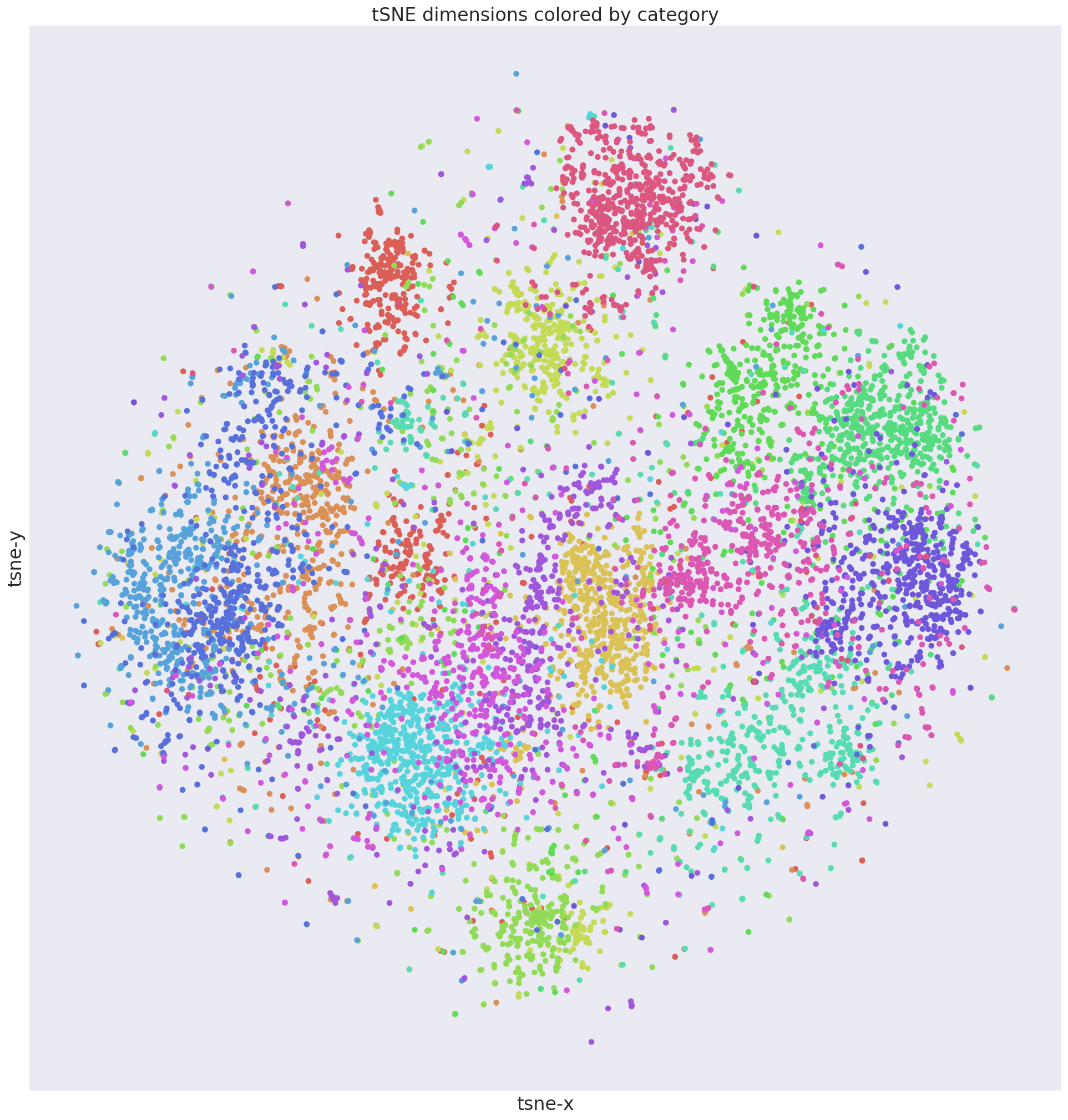}\\
        \vspace{-15pt}                
        \caption{\small{\textbf{Cosine-similarity based features of novel categories}}}
        \end{subfigure} 
        \hspace{0.001cm} 
        \begin{subfigure}[b]{0.49\textwidth}
        \includegraphics[width=\textwidth]{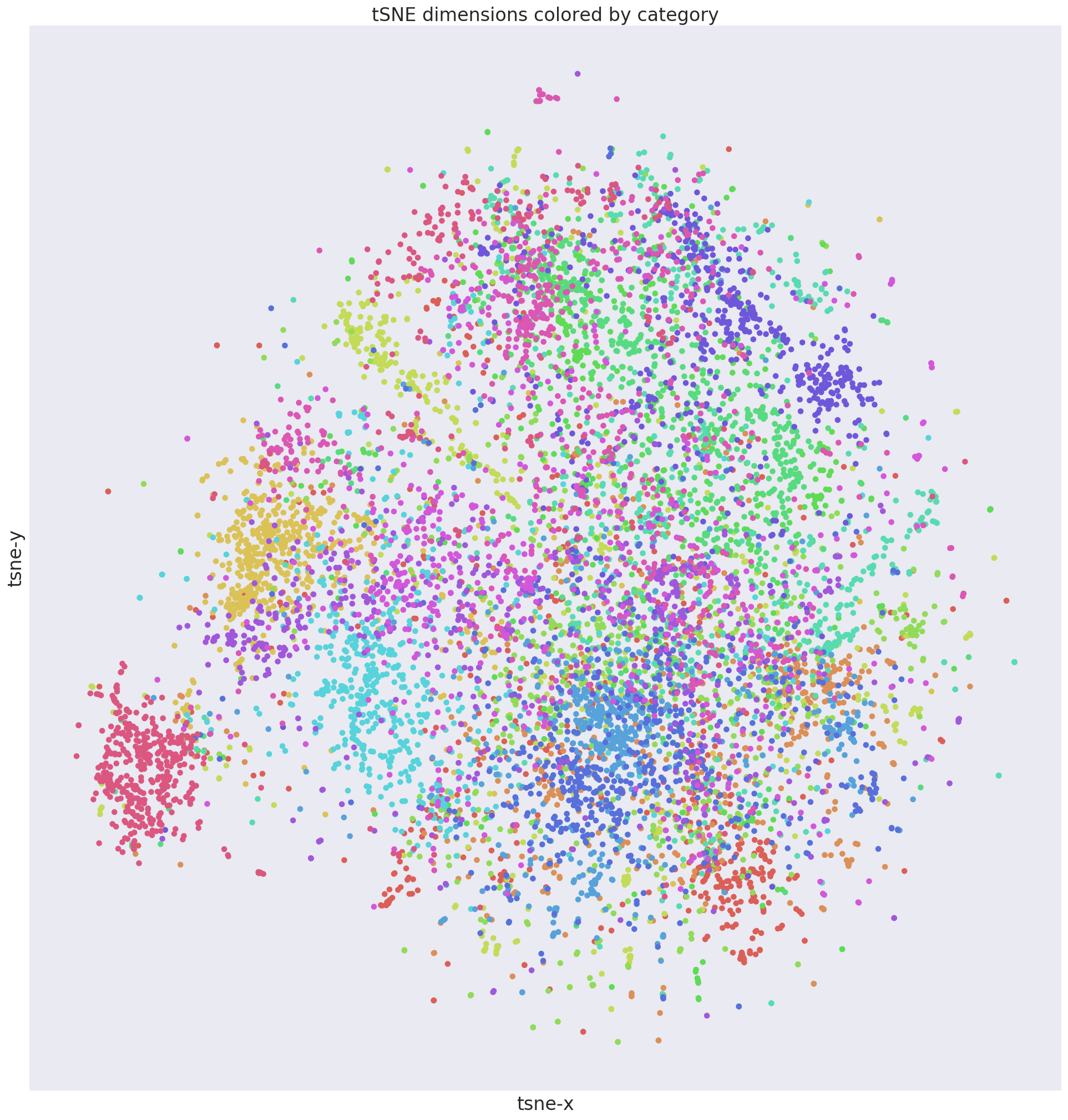}\\
        \vspace{-15pt}      
        \caption{\small{\textbf{Dot-product based features of novel categories}}}
        \end{subfigure}      
        \end{center}
        \vspace{5pt}
\end{subfigure}
\caption{\small{Here we visualize the t-SNE~\cite{maaten2008visualizing} scatter plots of the feature representations learned with \textbf{(a)} the cosine-similarity based ConvNet recognition model, and \textbf{(b)} the dot-product based ConvNet recognition model. 
Note that in the case of the cosine-similarity based ConvNet recognition model, we visualize the $l_2$-normalized features.
The visualized feature data points are from the ``unseen" during training validation categories of Mini-ImageNet. 
Each data point in the t-SNE scatter plots is colored according to its category.
}}
\label{fig:tsne}  
\end{figure*}

In addition to the above modification, we also choose to remove the ReLU non-linearity~\cite{nair2010rectified} after the last hidden layer of the feature extractor, which allows the feature vector $z$ to take both positive and negative values, similar to the classification weight vectors.
Note that the removal of the ReLU non-linearity does not make the composition of the last hidden layer with the classification layer a linear operation, since we $l_2$-normalize the feature vectors, which is a non-linear operation.
In our initial experiments with the cosine similarity based classifier we found that such a modification can significantly improve the recognition performance of novel categories.

We note that, although cosine similarity is already well established as an effective similarity function for classifying a test feature by comparing it with the available training features vectors~\cite{vinyals2016matching, mensink2012metric, rebuffi2017icarl}, 
in this work we use it for a different purpose, i.e., to replace the dot-product operation of the last linear layer of classification ConvNets used for applying the learnable weights of that layer to the test feature vectors.
The proposed modification in the architecture of a classification ConvNet allows to unify the recognition of base and novel categories without significantly altering the classification pipeline for the recognition of base categories (in contrast to~\cite{mensink2012metric, rebuffi2017icarl}).
To the best of our knowledge, employing the cosine similarity operation in such a way is novel in the context of few shot learning.
Interestingly, concurrently to us, Qi \etal~\cite{qi2017learning} also propose to use the cosine similarity function in a similar way for the few-shot learning task.
In a different line of work, 
very recently
Chunjie \etal~\cite{chunjie2017cosine} also explored cosine similarity for the typical supervised classification task.

\textbf{Advantages of cosine-similarity based classifier.}
Apart from making possible the unified recognition of both base and novel categories, \emph{the cosine-similarity based classifier leads the feature extractor to learn features that generalize significantly better on novel categories than features learned with the dot-product based classifier}.
A possible explanation for this is that, 
in order to minimize the classification loss of a cosine-similarity based ConvNet model, the $l_2$-normalized feature vector of an image must be very closely matched with the $l_2$-normalized classification weight vector of its ground truth category. 
As a consequence, the feature extractor is forced to (a) learn to encode on its feature activations exactly those discriminative visual cues that also the classification weight vectors of the ground truth categories learn to look for, and (b) learn to generate $l_2$-normalized feature vectors with low intraclass variance, since all the feature vectors that belong to the same category must be very closely matched with the single classification weight vector of that category.
This is visually illustrated in Figure~\ref{fig:tsne}, where we visualize t-SNE scatter plots of cosine-similarity-based and dot-product-based features related to categories ``unseen" during training. As can be clearly observed, the features generated from the cosine-similarity-based ConvNet form more compact and distinctive category-specific clusters (i.e., they provide more discriminative features).
Moreover, 
our cosine-similarity based classification objective
resembles the training objectives typically used by metric learning approaches~\cite{hoffer2015deep}.
In fact, it turns out that
\emph{
our feature extractor trained solely on cosine-similarity based classification of base categories, when used for image matching, it manages to surpass all prior state-of-the-art approaches on the few-shot object recognition task}.
 
\subsection{Few-shot classification weight generator} \label{sec:generator}

The few-shot classification weight generator $G(.,.|\phi)$ gets as input the feature vectors  $Z' = \{ z'_{i} \}_{i=1}^{N'}$ of the $N'$ training examples of a novel category (typically $N' \le 5$) and (optionally) the classification weight vectors of the base categories $W_{base}$. Based on them, it infers a classification weight vector $w' = G(Z',W_{base}|\phi)$ for that novel category. Here we explain how the above few-shot classification weight generator is constructed.

\textbf{Feature averaging based weight inference.}
Since, as we explained in section~\S~\ref{sec:classifier}, the cosine similarity based classifier of the ConvNet model forces the feature extractor to learn feature vectors that form compact category-wise clusters and the classification weight vectors to learn to be representative feature vectors of those clusters, an obvious choice is to infer the classification weight vector $w'$ 
by averaging the feature vectors of the training examples (after they have been $l_2$-normalized):
\begin{equation} \label{eq:wavg}
w'_{avg} = \frac{1}{N'} \sum_{i=1}^{N'} \overline{z}'_{i} \text{ .}
\end{equation}
The final classification weight vector in case we only use the feature averaging mechanism is: 
\begin{equation} \label{eq:wavgfinal}
w' = \phi_{avg} \odot w'_{avg} \text{ ,}
\end{equation}
where $\odot$ is the Hadamard product, and $\phi_{avg} \in \mathbb{R}^{d}$ is a learnable weight vector. Similar strategy has been previously proposed by Snell \etal~\cite{snell2017prototypical} 
and has demonstrated very good results.
However,
it does not fully exploit the 
knowledge about the visual world that the ConvNet model acquires during its training phase.
Furthermore, in case there is only a single training example for the novel category, the averaging cannot infer an accurate classification weight vector.

\textbf{Attention-based weight inference.}
We enhance the above feature averaging mechanism with an attention based mechanism that composes novel classification weight vectors by ``looking" at a memory that contains the base classification weight vectors $W_{base} = \{ w_b \}_{b=1}^{K_{base}}$. 
More specifically, an extra attention-based classification weight vector $w'_{att}$ is computed as:
\begin{equation} \label{eq:watt}
w'_{att} = \frac{1}{N'} \sum_{i=1}^{N'} \sum_{b=1}^{K_{base}} Att( \phi_{q} \overline{z}'_{i}, k_{b}) \cdot \overline{w}_{b} \text{ ,}
\end{equation}
where $\phi_{q} \in \mathbb{R}^{d \times d}$ is a learnable weight matrix that transforms the feature vector $\overline{z}'_{i}$ to query vector used for querying the memory, $\{ k_b \in \mathbb{R}^{d}\}_{b}^{K_{base}}$ is a set of $K_{base}$ learnable keys (one per base category) used for indexing the memory, and $Att(.,.)$ is an attention kernel implemented as a cosine similarity function\footnote{The cosine similarity scores are also scaled by a learnable scalar parameter $\gamma$ in order to increase the peakiness of the softmax distribution.} followed by a softmax operation over the $K_{base}$ base categories.
The final classification weight vector 
is computed as a weighted sum of the average based classification vector $w'_{avg}$ and the attention based classification vector $w'_{att}$:
\begin{equation} \label{eq:wattfinal}
w' = \phi_{avg} \odot w'_{avg} + \phi_{att} \odot w'_{att} \text{ ,}
\end{equation}
where $\odot$ is the Hadamard product, and $\phi_{avg}$, $\phi_{att} \in \mathbb{R}^{d}$ are learnable weight vectors. 

\textbf{Why using an attention-based weight composition?}
Thanks to the cosine-similarity based classifier, the base classification weight vectors learn to be representative feature vectors of their categories. 
Thus, the base classification weight vectors also encode visual similarity,
e.g., the classification vector of a mammal animal should be closer to the classification vector of another mammal animal rather than the classification vector of a vehicle. 
Therefore, the classification weight vector of a novel category can be composed as a linear combination of those base classification weight vectors that are most similar to the few training examples of that category. 
This allows our few-shot weight generator to explicitly exploit the acquired knowledge about the visual word (here represented by the base classification weight vectors) in order to improve the few-shot recognition performance.
This improvement is very significant especially in the one-shot recognition setting where averaging cannot provide an accurate classification weight vector.

\subsection{Training procedure}
\label{sec:training}
In order to learn the ConvNet-based recognition model 
(i.e. the feature extractor $F(.|\theta)$ as well as the classifier $C(.|W^*)$) 
and the few-shot classification weight generator $G(.,.|\phi)$, we use as 
the sole input a training set $D_{train} = \bigcup_{b=1}^{K_{base}} \{ x_{b,i} \}_{i=1}^{N_b}$ of $K_{base}$ base categories. We split the training procedure into 2 stages and at each stage we minimize a different cross-entropy loss of the following form:
\begin{equation} \label{eq:loss}
\frac{1}{K_{base}} \sum_{b=1}^{K_{base}} \frac{1}{N_b} \sum_{i=1}^{N_b} loss( x_{b,i}, b ),
\end{equation}
where $loss(x, y)$ is the negative log-probability $-log( p_y )$ of the $y$-th category in the probability vector $p = C( F(x | \theta ) | W^*) $. 
The meaning of $W^*$ is different on each of the training stages, as we explain below.

\textbf{1st training stage:} During this stage we only learn the ConvNet recognition model without the few-shot classification weight generator. 
Specifically, at this stage we learn the parameters $\theta$ of the feature extractor $F(.|\theta)$ and the base classification weight vectors $W_{base} = \{w_b\}_{b=1}^{K_{base}}$. This is done in exactly the same way as for any other standard recognition model.
In this case $W^*$ is equal to the base classification weight vectors $W_{base}$.

\textbf{2nd training stage:} 
During this stage we train the learnable parameters $\phi$ of the few-shot classification weight generator while we continue training the base classification weight vectors $W_{base}$ (in our experiments during that training stage we freezed the feature extractor).
In order to train the few-show classification weight generator, in each batch we randomly pick $K_{novel}$ ``fake" novel categories from the base categories and we treat them in the same way as we will treat the actual novel categories after training. 
Specifically, instead of using the classification weight vectors in $W_{base}$ for those ``fake" novel categories, 
we sample $N'$ training examples (typically $N' \le 5$) for each of them, compute their feature vectors $Z'=\{ z'_i \}_{i=1}^{N'}$, and give those feature vectors to the few-shot classification weight generator $G(.,.|\phi)$ in order to compute novel classification weight generators. The inferred classification weight vectors are used for recognizing the ``fake" novel categories. Everything is trained end-to-end. Note that we take care to exclude from the base classification weight vectors that are given as a second argument to the few-shot weight generator $G(.,.|\phi)$ those classification vectors that correspond to the ``fake" novel categories.
In this case $W^*$ is the union of the ``fake" novel classification weight vectors generated by $G(.,.|\phi)$ and the classification weight vectors of the remaining base categories. More implementation details of this training stage are provided in appendix~\ref{sec:tdetails}.

\section{Experimental results} \label{sec:experimentalresults}

We extensively evaluate the proposed few-shot recognition system w.r.t. both its few-shot recognition performance of novel categories and its ability to not ``forget" the base categories on which it was trained. 

\subsection{Mini-ImageNet experiments} \label{sec:miniI}
\begin{table*}[t]
\centering
\renewcommand{\figurename}{Table}
\renewcommand{\captionlabelfont}{\bf}
\renewcommand{\captionfont}{\small}
\resizebox{0.75\textwidth}{!}{
{\setlength{\extrarowheight}{2pt}\scriptsize
{
\hspace{-10pt}
\begin{tabular}{l <{\hspace{-0.3em}} | >{\hspace{-0.5em}} r >{\hspace{-0.5em}} r >{\hspace{-0.5em}} r | >{\hspace{-0.5em}} r >{\hspace{-0.5em}} r >{\hspace{-0.5em}} r}
\toprule
\multicolumn{1}{l|}{\multirow{ 2}{*}{Models} } & \multicolumn{3}{c|}{5-Shot learning -- $K_{novel}$=5} & \multicolumn{3}{c}{1-Shot learning -- $K_{novel}$=5}\\
& Novel & Base & Both & Novel & Base & Both\\
\midrule
\;Matching-Nets~\cite{vinyals2016matching} 	   & 68.87 $\pm$ 0.38\% & - & - & 55.53 $\pm$ 0.48\% & - & -\\
\;Prototypical-Nets~\cite{snell2017prototypical}  & 72.67 $\pm$ 0.37\%  & 62.10\% & 32.70\% & 54.44 $\pm$ 0.48\%  & 52.35\% & 26.68\%\\
\midrule
\emph{Ours} & & & & & \\
\;Cosine Classifier & 72.83 $\pm$ 0.35\%  & 70.68\% & 51.89\% & 54.55 $\pm$ 0.44\%  & 70.68\% & 39.17\%\\
\;Cosine Classifier \& Avg. Weight Gen & 74.66 $\pm$ 0.35\%  & 70.92\% & 60.26\% & 55.33 $\pm$ 0.46\%  & 70.45\% & 48.56\%\\
\;Cosine Classifier \& Att. Weight Gen & \textbf{74.92 $\pm$ 0.36\%}  & 70.88\% & 60.50\%  & \textbf{58.55 $\pm$ 0.50}\%  & 
70.73\% & 50.50\%\\
\midrule
\emph{Ablations} & & & & & \\
\;Dot Product & 64.58 $\pm$ 0.38\% & 63.59\% & 31.80\%   & 46.09 $\pm$ 0.40\%  & 63.59\% & 24.76\%\\
\;Dot Product \& Avg. Weight Gen & 60.30 $\pm$ 0.39\% & 62.15\% & 46.41\%   & 44.31 $\pm$ 0.40\%  & 61.99\% & 39.05\%\\
\;Dot Product \& Att. Weight Gen & 67.81 $\pm$ 0.37\% & 62.11\% & 48.70\%   & 53.88 $\pm$ 0.48\%  & 62.28\% & 42.41\%\\
\midrule
\emph{Ablations} & & & & & \\
\;Cosine w/ ReLU. & 71.04 $\pm$ 0.36\%  & \textbf{72.51\%} & 58.16\% & 52.91 $\pm$ 0.45\%  & \textbf{72.51\%} & 43.17\%\\
\;Cosine w/ ReLU. \& Avg. Weight Gen & 71.30 $\pm$ 0.38\%  & 72.47\% & 59.33\% & 53.19 $\pm$ 0.45\%  & 71.70\% & 49.53\%\\
\;Cosine w/ ReLU. \& Att. Weight Gen & 73.03 $\pm$ 0.38\%  & 72.26\% & \textbf{61.05\%} & 56.09 $\pm$ 0.54\%  & 72.34\% & \textbf{51.25\%}\\
\bottomrule
\end{tabular}}}}
\vspace{3pt}
\caption{
\small{Average classification accuracies on the validation set of Mini-ImageNet. 
The Novel columns report the average 5-way and 1-shot or 5-shot classification accuracies of novel categories (with $95\%$ confidence intervals), 
the Base and Both columns report the classification accuracies of base categories and of both type of categories respectively.
In order to report those results we sampled 2000 tasks each with $15 \times 5$ test examples of novel categories and $15 \times 5$ test examples of base categories.}}
\label{tab:miniINval}
\end{table*} 
\begin{table*}[t]
\centering
\renewcommand{\figurename}{Table}
\renewcommand{\captionlabelfont}{\bf}
\renewcommand{\captionfont}{\small} 
\resizebox{0.75\textwidth}{!}{
{\setlength{\extrarowheight}{2pt}\scriptsize
{
\hspace{-10pt}
\begin{tabular}{l <{\hspace{-0.3em}} | >{\hspace{-0.5em}} r | >{\hspace{-0.5em}} r >{\hspace{-0.5em}} r >{\hspace{-0.5em}} r | >{\hspace{-0.5em}} r >{\hspace{-0.5em}} r >{\hspace{-0.5em}} r}
\toprule
\multicolumn{1}{l|}{\multirow{ 2}{*}{Models} } & \multicolumn{1}{l|}{\multirow{ 2}{0.9cm}{Feature Extractor} } & \multicolumn{3}{c|}{5-Shot learning -- $K_{novel}$=5} & \multicolumn{3}{c}{1-Shot learning -- $K_{novel}$=5}\\
& & Novel & Base & Both & Novel & Base & Both\\
\midrule
Matching-Nets~\cite{vinyals2016matching} & \abbr{C64F} & 55.30\% & - & -   & 43.60\% 		   & - & -\\
Ravi and Laroche~\cite{ravi2016optimization} & \abbr{C32F} & 60.20 $\pm$ 0.71\% & - & - & 43.40 $\pm$ 0.77\% & - & -\\
Finn \etal~\cite{finn2017model} & \abbr{C64F} & 63.10 $\pm$ 0.92\% & - & - & 48.70 $\pm$ 1.84\% & - & -\\
Prototypical-Nets~\cite{snell2017prototypical} & \abbr{C64F} & 68.20 $\pm$ 0.66\% & - & - & 49.42 $\pm$ 0.78\% & - & -\\
Mishra \etal~\cite{mishra2017meta} & \abbr{ResNet} & 68.88 $\pm$ 0.92\% & - & - & 55.71 $\pm$ 0.99\% & - & -\\
\midrule
Ours & \abbr{C32F}  & 70.27 $\pm$ 0.64\%  & 61.08\% & 52.45\% & 54.33 $\pm$ 0.81\%  & 61.09\% & 43.05\%\\
Ours & \abbr{C64F}  & 72.81 $\pm$ 0.62\%  & 68.13\% & 57.72\% & \textbf{56.20 $\pm$ 0.86\%}  & 68.08\% & 48.09\%\\
Ours & \abbr{C128F} & \textbf{73.00 $\pm$ 0.64\%} & 70.90\% & \textbf{59.35\%} & 55.95 $\pm$ 0.84\%  & 70.72\% & 49.08\%\\
Ours & \abbr{ResNet} & 70.13 $\pm$ 0.68\% & \textbf{80.16\%} & 56.04\% & 55.45 $\pm$ 0.89\% & \textbf{80.24\%} & \textbf{51.23\%}\\
\bottomrule
\end{tabular}}}}
\vspace{3pt}
\caption{
\small{Average classification accuracies on the test set of Mini-ImageNet. 
In order to report those results we sampled 600 tasks in a similar fashion as for the validation set of Mini-ImageNet.
}}
\label{tab:miniINtest}
\end{table*}

\textbf{Evaluation setting for recognition of novel categories.}
We evaluate our few-shot object recognition system on the Mini-ImageNet dataset~\cite{vinyals2016matching} that
includes 100 different categories with 600 images per category, each of size $84 \times 84$.
For our experiments we used the splits by Ravi and Laroche~\cite{ravi2016optimization} that include 64 categories for training, 16 categories for validation, and 20 categories for testing.
The typical evaluation setting on this dataset is first to train 
a few-shot model on the training categories and then during test time to use the validation (or the test) categories in order to form few-shot tasks on which the trained model is evaluated.
Those few-shot tasks are formed by first sampling $K_{novel}$ categories and one or five training example per category (1-shot and 5-shot settings respectively), which the trained model uses for meta-learning those categories, and then evaluating it on some test examples that come from the same novel categories but do not overlap with the training examples.

\textbf{Evaluation setting for the recognition of the base categories.}
When we evaluate our model w.r.t. few-shot recognition task on the validation / test categories, 
we consider as base categories the $64$ training categories on which we trained the model.
Since the proposed few-shot object recognition system has the ability to not forget the base categories, we would like to also evaluate the recognition performance of our model on those base categories.
Therefore, we sampled 300 extra images for each training category that we use as validation image set 
for the evaluation of the recognition performance of the base categories and also another 300 extra images that are used for the same reason as test image set.
Therefore, when we evaluate our model w.r.t. the few-shot learning task on the validation / test categories we also evaluate w.r.t. recognition performance of the base categories on the validation / test image set of the training categories.

\subsubsection{Ablation study} \label{sec:ablation}

In Table~\ref{tab:miniINval} we provide an ablation study of the proposed object recognition framework on the validation set of mini-ImageNet. 
We also compare with two prior state-of-the-art approaches, 
Prototypical Networks~\cite{snell2017prototypical} and Matching Nets~\cite{vinyals2016matching},
that we re-implemented ourselves in order to ensure a fair comparison.
The feature extractor used in all cases is a ConvNet model that has 4 convolutional modules,
with $3 \times 3$ convolutions, followed by batch normalization, ReLU nonlinearity\footnote{Unless otherwise stated, our cosine-similarity based models as well as the re-implementation of Matching-Nets do not have a ReLU nonlinearity after the last convolutional layer, since in both cases this modification improved the recognition performance on the few-shot recognition task}, and $2 \times 2$ max-pooling. 
Given as input images of size $84 \times 84$ it yields feature maps with spatial size $5 \times 5$. The first two convolutional layers have $64$ feature channels and the latter two have $128$ feature channels.

\textbf{Cosine-similarity based ConvNet model.}
First we examine the performance of the cosine-similarity based ConvNet recognition model (entry Cosine Classifier) without training the few-shot classification weight generator (i.e., we only perform the 1st training stage as was described in section~\ref{sec:training}).
In order to test its performance on the novel categories, during test time we estimate classification weight vectors using feature averaging.
\emph{We want to stress out that in this case there are no learnable parameters involved in the generation of the novel classification weight vectors and also the ConvNet model it was never trained on the few-shot recognition task.
Despite that, the features learned by the cosine-similarity based ConvNet model matches or even surpasses the performance of the Matching-Nets and Prototypical Networks, which are explicitly trained on the few-shot object recognition task.} 
By comparing the cosine-similarity based ConvNet models (Cosine Classifier entries) with the dot-product based models (Dot Product entries) we observe that the former drastically improve the few-shot object recognition performance, which means that the feature extractor that is learned with the cosine-similarity classifier generalizes significantly better on ``unseen" categories than the feature extractor learned with the dot-product classifier.
Notably, the cosine-similarity classifier significantly improves also the recognition performance on the base categories.

\textbf{Removing the last ReLU unit.}
In our work we propose to remove the last ReLU non-linearity from the feature extractor when using a cosine classifier.
Instead, keeping the ReLU units (Cosine w/ ReLU entries) decreases the accuracy on novel categories while increasing it on base categories.

\textbf{Few-shot classification weight generator.}
Here we examine the performance of our system when we also incorporate on it the proposed few-shot classification weight generator. 
In Table~\ref{tab:miniINval} we provide two solutions for the few-shot weight generator: the entry Cosine Classifier \& Avg. Weight Gen that uses only the feature averaging mechanism described in section~\ref{sec:generator} and the entry Cosine Classifier \& Att. Weight Gen that uses both the feature averaging and the attention based mechanism.
Both types of few-shot weight generators are trained during the 2nd training stage that is described in section~\ref{sec:training}. 
We observe that both of them offer a very significant boost on the few-shot recognition performance of the cosine similarity based model (entry Cosine Classifier). 
Among the two, the attention based solution exhibits better few-shot recognition behavior, especially in the 1-shot setting where it has more than 3 percentage points higher performance.
Also, it is easy to see that the few-shot classification weight generator does not affect the recognition performance of the base categories, which is around $70.50\%$ in all the cosine-similarity based models.
Moreover, by introducing the few-shot weight generator, the recognition performance in both type of categories (columns Both) increases significantly, 
which means that the ConvNet model achieves better behavior w.r.t. our goal of unified recognition of both base and novel categories.
The few-shot recognition performance of our full system, which is the one that includes the attention based few-shot weight generator (entry Cosine classifier \& Att. Weight Gen), offers a very significant improvement w.r.t. the prior state-of-the-art approaches on the few-shot object recognition task, i.e., from $72.67\%$ to $74.92\%$ in the 5-shot setting and from $55.53\%$ to $58.55\%$ in the 1-shot setting.
Also, our system achieves significantly higher performance on the recognition of base categories compared to Prototypical Networks\footnote{In order to recognize base categories with Prototypical Networks, the prototypes for the base categories are computed by averaging all the available training features vectors}.

\subsubsection{Comparison with state-of-the-art}
\begin{table*}[t!]
\centering
\renewcommand{\figurename}{Table}
\renewcommand{\captionlabelfont}{\bf}
\renewcommand{\captionfont}{\small} 
\renewcommand{\arraystretch}{1.2}
\renewcommand{\tabcolsep}{1.2mm}
\resizebox{0.95\linewidth}{!}{
\begin{tabular}{l | ccccc@{\hspace{8mm}}ccccc@{\hspace{8mm}}ccccc@{\hspace{8mm}}}
& \multicolumn{5}{c@{\hspace{8mm}}}{Novel} & \multicolumn{5}{c@{\hspace{8mm}}}{All} & \multicolumn{5}{c@{\hspace{8mm}}}{All with prior}\\
Approach &$N'$=1 & 2 & 5 & 10 & 20&$N'$=1 & 2 & 5 & 10 & 20&$N'$=1 & 2 & 5 & 10 & 20\\
\midrule
\emph{Prior work} &  &  &  &  &  &  &  &  &  &  &  &  &  & \\
\;Prototypical-Nets~\cite{snell2017prototypical} (from~\cite{wang2018low}) & 39.3 & 54.4 & 66.3 & 71.2 & 73.9 & 49.5 & 61.0 & 69.7 & 72.9 & 74.6 & 53.6 & 61.4 & 68.8 & 72.0 & 73.8\\
\;Matching Networks~\cite{vinyals2016matching} (from~\cite{wang2018low}) & 43.6 & 54.0 & 66.0 & 72.5 & 76.9 & 54.4 & 61.0 & 69.0 & 73.7 & 76.5 & 54.5 & 60.7 & 68.2 & 72.6 & 75.6\\
\;Logistic regression (from~\cite{wang2018low}) & 38.4 & 51.1 & 64.8 & 71.6 & 76.6 & 40.8 & 49.9 & 64.2 & 71.9 & 76.9 & 52.9 & 60.4 & 68.6 & 72.9 & 76.3\\
\;Logistic regression w/ H~\cite{hariharan2016low} (from~\cite{wang2018low}) & 40.7 & 50.8 & 62.0 & 69.3 & 76.5 & 52.2 & 59.4 & 67.6 & 72.8 & 76.9 & 53.2 & 59.1 & 66.8 & 71.7 & 76.3\\
\;SGM w/ H~\cite{hariharan2016low}	& - & - & - & - & - & 54.3 & 62.1 & 71.3 & 75.8 & 78.1 & - & - & - & - & -\\
\;Batch SGM~\cite{hariharan2016low} & - & - & - & - & - & 49.3 & 60.5 & 71.4 & 75.8 & 78.5 & - & - & - & - & - \\
\midrule
\emph{Concurrent work} &  &  &  &  &  &  &  &  &  &  &  &  &  & \\
\;Prototype Matching Nets w/ H~\cite{wang2018low} & 45.8 & \textbf{57.8} & 69.0 & 74.3 & 77.4 & 57.6 & 64.7 & 71.9 & 75.2 & 77.5 & 56.4 & 63.3 & 70.6 & 74.0 & 76.2\\
\;Prototype Matching Nets~\cite{wang2018low} & 43.3 & 55.7 & 68.4 & 74.0 & 77.0 & 55.8 & 63.1 & 71.1 & 75.0 & 77.1 & 54.7 & 62.0 & 70.2 & 73.9 & 75.9\\
\midrule
\emph{Ours} &  &  &  &  &  &  &  &  &  &  &  &  &  & \\
\multirow{ 1}{*}{\;Cosine Classifier \& Avg. Weight Gen} & 45.23 & 56.90 & 68.68 & 74.36 & 77.69 & 57.65 & 64.69 & 72.35 & 76.18 & 78.46 & 56.43 & 63.41 & 70.95 & 74.75 & 77.00\\
& $\pm$ .25 & $\pm$ .16 & $\pm$ .09 & $\pm$ .06 & $\pm$ .06 & $\pm$ .15 & $\pm$ .10 & $\pm$ .06 & $\pm$ .04 & $\pm$ .04 & $\pm$ .15 & $\pm$ .10 & $\pm$ .06 & $\pm$ .04 & $\pm$ .03\\
\multirow{ 1}{*}{\;Cosine Classifier \& Att. Weight Gen} & \textbf{46.02} & 57.51 & \textbf{69.16} & \textbf{74.83} & \textbf{78.11} & \textbf{58.16} & \textbf{65.21} & \textbf{72.72} & \textbf{76.50} & \textbf{78.74} & \textbf{56.76} & \textbf{63.80} & \textbf{72.72} & \textbf{75.02} & \textbf{77.25}\\
& $\pm$ \textbf{.25} & $\pm$ .15 & $\pm$ \textbf{.09} & $\pm$ \textbf{.06} & $\pm$ \textbf{.05} & $\pm$ \textbf{.15} & $\pm$ \textbf{.09} & $\pm$ \textbf{.06} & $\pm$ \textbf{.04} & $\pm$ \textbf{.03} & $\pm$ \textbf{.15} & $\pm$ \textbf{.10} & $\pm$ \textbf{.06} & $\pm$ \textbf{.04} & $\pm$ \textbf{.04}\\
\midrule
\end{tabular}
}
\caption{
Top-5 accuracy on the novel categories and on all categories (with and without priors) fot the ImageNet based few-shot benchmark proposed in~\cite{hariharan2016low} (for more details about the evaluation metrics we refer to~\cite{wang2018low}).
For each novel category we use $N'=$ 1, 2, 5, 10 or 20 training examples.
Methods with ``w/ H" use mechanisms that hallucinate extra training examples for the novel categories.
The second rows in our entries report the $95\%$ confidence intervals.
}
\label{tab:everything}
\end{table*}

Here we compare the proposed few-shot object recognition system with other state-of-the-art approaches on the Mini-ImageNet test set. 

\textbf{Explored feature extractor architectures.}
Because prior approaches use several different network architectures for implementing the feature extractor of the ConvNet model, we evaluate our model with each of those architectures. 
Specifically the architectures that we evaluated are:
\abbr{C32F} is a 4 module ConvNet network (which was described in \S~\ref{sec:ablation}) with 32 feature channels on each convolutional layer, 
\abbr{C64F} has 64 feature channels on each layer, and
in \abbr{C128F} the first two layers have 64 channels and the latter two have 128 channels (exactly the same as the model that was used in \S~\ref{sec:ablation}).
With \abbr{ResNet} we refer to the ResNet~\cite{he2016deep} like network that was used from Mishra \etal~\cite{mishra2017meta} (for more details we refer to~\cite{mishra2017meta}). 

In Table~\ref{tab:miniINtest} we provide the experimental results. 
In all cases, our models (that include the cosine-similarity based ConvNet model and the attention-based few-shot weight generator)   
achieve better few-shot object recognition performance than prior approaches.
\emph{Moreover, it is very important to note that our approach is capable to achieve such excellent accuracy on the novel categories while at the same time 
it does not sacrifice the recognition performance of the base categories,
which is an ability that prior methods lack.}

\subsubsection{Qualitative evaluation with t-SNE scatter plots} \label{sec:tsne}

Here we compare qualitatively the feature representations learned by the proposed cosine-similarity based ConvNet recognition model with those learned by the typical dot-product based ConvNet recognition model. 
For that purpose in Figure~\ref{fig:tsne} we provide the t-SNE~\cite{maaten2008visualizing} scatter plots that visualize the local-structures of the feature representations learned in those two cases. 
Note that the visualized features are from the validation categories of the Mini-ImageNet dataset that are ``unseen" during training.
Also, in the case of the cosine-similarity based ConvNet recognition model, we visualize the $l_2$-normalized features, 
which are the features that are actually learned by the feature extractor.

We observe that 
the feature extractor learned with the cosine-similarity based ConvNet recognition model, 
when applied on the images of ``unseen" categories (in this case the validation categories of Mini-ImageNet), 
it generates features that form more compact and distinctive category-specific clusters (i.e., more discriminative features).
Due to that, as it was argued in section~\S\ref{sec:classifier}, 
the features learned with the proposed cosine-similarity based recognition model generalize better on the ``unseen" categories than the features learned with the typical dot-product based recognition model.

\subsection{Few-shot benchmark of Bharath \& Girshick~\cite{hariharan2016low}}

Here we evaluate our approach on the ImageNet based few-shot benchmark proposed by Bharath and Girshick~\cite{hariharan2016low}
using the improved evaluation metrics proposed by Wang \etal~\cite{wang2018low}.
Briefly, this benchmark splits the ImageNet categories into 389 base categories and 611 novel categories; 
193 of the base categories and 300 of the novel categories are used for cross validation
and the remaining 196 base categories and 311 novel categories are used for the final evaluation (for more details we refer to~\cite{hariharan2016low}).
We use the same categories split as they did. 
However, because it was not possible to use the same training images that they did for the novel categories\footnote{It was not possible to establish a correspondence between the index files that they provide and the ImageNet images}, 
we sample ourselves $N'$ training images per novel category and, similar to them, evaluate using the images in the validation set of ImageNet.
We repeat the above experiment 100 times (sampling each time a different set of training images for the novel categories) and report in Table~\ref{tab:everything} the mean accuracies and the $95\%$ confidence intervals for the recognition accuracy metrics proposed in~\cite{wang2018low}. 

\textbf{Comparison to prior and concurrent work.}
We compare our full system (Cosine Classifier \& Att. Weight Gen entry) against prior work, such as
Prototypical-Nets~\cite{snell2017prototypical},
Matching Networks~\cite{vinyals2016matching},
and the work of Bharath and Girshick~\cite{hariharan2016low}.
We also compare against the work of Wang \etal~\cite{wang2018low}, which is concurrent to ours.
We observe that in all cases our approach achieves superior performance than prior approaches and even exceeds (in all but one cases) the Prototype Matching Net~\cite{wang2018low} based approaches that are concurrent to our work.

\textbf{Feature extractor:}
The feature extractor of all approaches is implemented with a \emph{ResNet-10}~\cite{he2016deep} network architecture\footnote{Similar to what it is already explained, our model does not include the last ReLU non-linearity of the \emph{ResNet-10} feature extractor} that gets as input images of $224 \times 224$ size.
Also, when training the attention based few-shot classification weight generator component of our model (2nd training stage) we found helpful to apply dropout with $0.5$ probability on the feature vectors generated by the feature extractor.

\section{Conclusions} \label{sec:conclusion}

In our work we propose a dynamic few-shot object recognition system that is able to quickly learn novel categories without forgetting the base categories on which it was trained, a property that most prior approaches on the few-shot learning task neglect to fulfill.
To achieve that goal we propose a novel attention based few-shot classification weight generator 
as well as a cosine-similarity based ConvNet classifier.
This allows to recognize in a unified way both novel and base categories and also leads to learn feature representations with better generalization capabilities.
We evaluate our framework on Mini-ImageNet
and the recently introduced fews-shot benchmark of Bharath and Girshick~\cite{hariharan2016low}
where we demonstrate that our approach is capable of both maintaining high recognition accuracy on base categories and to achieve excellent few-shot recognition accuracy on novel categories that surpasses prior state-of-the-art approaches by a significant margin. 

{\small
\bibliographystyle{ieee}
\bibliography{egbib}
}

\begin{appendices}

\section{Implementation details of training procedure followed during the 2nd training stage} \label{sec:tdetails}

As explained in~\S\ref{sec:training}, in order to train the few-shot classification weight generator, during the 2nd training stage we sample $K_{novel}$ ``fake" novel categories from the base training categories and we treat them in the same way as we will treat the actual novel categories after training.
More specifically, during the 2nd training stage we form ``training episodes";
each ``training episode" is created by sampling:
\textbf{(a)} $K_{novel}$ ``fake" novel categories with $N'$ training examples per ``fake" novel category, 
\textbf{(b)} $T_{novel}$ test image examples from the ``fake" novel categories, and 
\textbf{(c)} $T_{base}$ test image examples from the remaining base categories (i.e., the base categories without the ``fake" novel categories). 
Given such a ``training episode", we first use the $N'$ training examples of each ``fake" novel category to infer with the few-shot weight generator a ``fake" novel classification weight vector for that category and
then we use the union of ``fake" novel classification weight vectors and the classification weight vectors of the remaining base
categories in order to learn to classify the $T=T_{novel}+T_{base}$ test image examples. 
To conclude, in order to train the few-shot classification weight generator and the classification weight vectors of the base categories we use stochastic gradient descent based routines with training batches that include multiple different instances of the above ``training episodes".

\end{appendices}

\end{document}